\documentclass[conference]{IEEEtran}
\IEEEoverridecommandlockouts
\usepackage{cite}
\usepackage{amsmath,amssymb,amsfonts}
\usepackage{subfigure}
\usepackage{graphicx}
\usepackage{textcomp}
\usepackage{xcolor}
\def\BibTeX{{\rm B\kern-.05em{\sc i\kern-.025em b}\kern-.08em
    T\kern-.1667em\lower.7ex\hbox{E}\kern-.125emX}}
\usepackage{booktabs} 
\usepackage{comment}
\usepackage{enumitem}
\usepackage{amsfonts,amssymb}

\usepackage{algorithm}
\usepackage{algpseudocode}
\usepackage{amsmath}
\begin{document}

\title{Are You A Risk Taker? Adversarial Learning of Asymmetric Cross-Domain Alignment for Risk Tolerance Prediction}

\author{Zhe Liu\textsuperscript{\rm 1}, Lina Yao\textsuperscript{\rm 1}, Xianzhi Wang\textsuperscript{\rm 2}, Lei Bai \textsuperscript{\rm 1},  Jake An\textsuperscript{\rm 3}\\
\textsuperscript{\rm 1}School of Computer Science and Engineering, University of New South Wales, Australia\\ 
\textsuperscript{\rm 2}School of Computer Science, University of Technology Sydney, Australia \\
\textsuperscript{\rm 3}Raiz Invest Limited, Australia \\
\textsuperscript{\rm 1}\{zhe.liu1,lei.bai\}@student.unsw.edu.au\\
\textsuperscript{\rm 1}\{lina.yao\}@unsw.edu.au\\
\textsuperscript{\rm 2}xianzhi.wang@uts.edu.au \\
\textsuperscript{\rm 3}jake@raizinvest.com.au
}
\maketitle

\begin{abstract}
Most current studies on survey analysis and risk tolerance modelling lack professional knowledge and domain-specific models. Given the effectiveness of generative adversarial learning in cross-domain information, we design an Asymmetric cross-Domain Generative Adversarial Network (ADGAN) for domain scale inequality. ADGAN utilizes the information-sufficient domain to provide extra information to improve the representation learning on the information-insufficient domain via domain alignment. We provide data analysis and user model on two data sources: Consumer Consumption Information and Survey Information. We further test ADGAN on a real-world dataset with view embedding structures and show ADGAN can better deal with the class imbalance and unqualified data space than state-of-the-art, demonstrating the effectiveness of leveraging asymmetrical domain information.
\end{abstract}

\section{Introduction}
E-finance has fundamentally changed the landscape and ways of information transfer in the finance industry~\cite{haddad2018finance}.
Consumers nowadays can easily access their assets and manage their investments via mobile devices.
Given the massiveness of consumers and intricate analysis work, finance analysts and agents are increasingly relying on machine learning to suggest suitable portfolios to consumers, where
Financial Risk Tolerance (FRT)\footnote{Financial risk tolerance refers to the degree of variability that the consumers can accept the negative changes in the value of investment or an outcome that is adversely different from the expected one~\cite{kannadhasan2016relationship}.} serves as a crucial measure to assess individuals' risk preferences on investment choices~\cite{nguyen2019joint}.
To accommodate consumers' risk expectations, the standard practice for recommending investment portfolios is to design questionnaires to survey consumers and then to assess their FRT. While survey results show significant correlations between consumers' feedbacks on the questionnaire and their real FRT levels~\cite{kannadhasan2016relationship,onsomu2017risk,martin2018examining}, consumers' feedbacks are liable to be biased by consumers investigated and inauthentic information.
Besides, consumers providing similar feedbacks may have differed FRT (i.e., the multivalued function phenomenon), and consumers' FRT often has an imbalanced class distribution.

Consumers' online activities, especially e-commerce transactions, are potentially an excellent source of information for assessing consumers' risk preferences.
However, existing researches on online activities focus on predicting consumers' future behaviors based on their past online activities~\cite{qiu2015predicting} while rarely considering the possible hidden relationship between the patterns and FRT.
Some studies suggest that shallow online activities such as \textit{clicks}, \textit{adding to cart}, and \textit{purchasing} could imply deeper consumers' behavior patterns on consumption habits and preferences~\cite{kooti2016portrait,su2018user}. Nevertheless, they have not considered domain knowledge to leverage the data fully.

Leveraging both consumer surveys and online activities for FRT would require aligning records from the two sources and learning unified presentations from data.
The above poses significant challenges to the current research.
First, there lack unified supervised methods to use non-pairwise domain data in a generative adversarial domain alignment.
Second, besides the challenges for each domain dataset, the data from diffident domains might be of different sizes and have small overlaps, meaning they are hardly aligned.
%
Given Generative Adversarial Network (GAN)~\cite{cgan,gan2, zhang2019adversarial}'s outstanding performance in generating samples, we consider it to be a promising solution to address the above challenges.
Although some previous studies have applied GAN for domain alignment~\cite{gan1,domainadaption1,compare2}, e.g.,  using records of the same object from different days to co-train the embedding layers~\cite{domainadaption1},
the existing studies fail to notice the scale asymmetry of multi-domain data and leverage this characteristic to enhance the representation learning of the insufficient domain in the unified model.
We make the following contributions in this paper:

\begin{itemize}[topsep=3pt]
    \item We analyze consumers' transactions (Section \ref{sec:Data Character}), consumption structures (Stratum Feature) and shopping affiliation (Life Feature) using a domain-specific consumer consumption model.
    
    \item
    We develop a novel ADGAN (Section~\ref{methodology}) to handle asymmetric multi-domain data. ADGAN employs two specially designed batch construction methods in a unified model for domain alignment and insufficient domain learning (Section~\ref{sec:survey_analysis}). It shows significant improvement over state-of-the-art algorithms.
    
    \item
   We design data-specific view embedding structures (Section~\ref{model_definition}) to extract information from raw data and a Gaussian noise fusion (Section~\ref{sef:traning}) to overcome the multivalued function phenomenon in survey data.
   \item Our model shows robustness in handling imbalanced cross-domain data. Our experimental results on a real-world dataset demonstrate its superiority to a series of baselines and state-of-the-art methods.

\end{itemize}
\section{Related Work}



\subsection{FRT-related Theories}

Business researchers show demographics, environmental factors, psychology, economics, and biosociology all influence consumers' judgments to finance-related decisions~\cite{kannadhasan2016relationship,tang2016self}.
They have designed factor-specific questionnaires to measure and analyze FRT. 
For example,
questionnaires designed from the philanthropy perspective revealed that consumers' corporate social responsibility would affect their purchase behaviors~\cite{lichtenstein2004effect,mohr2005effects};
an investigation of 726 consumers \cite{hartmann2012consumer} confirmed the positive correlation between environmental psychology with financial decisions;
the collectivism-individualism theory~\cite{triandis2018individualism}
suggested different types of people would make decisions driven by disparate deep reasons---some were motivated by personal preference and demands while others valued more about the entity benefits.
%
Roehrich et al.~\cite{roehrich2004consumer} indicated consumer was the main force for innovative behaviors.
%
The above business research concentrated on building a systematic consumer model with nameable attributes.
Such research can explain the implication of features but provide little help in designing related algorithms~\cite{nguyen2019joint}. Some studies~\cite{buskirk2018surveyintroduction} further apply linear regression and deep learning.
But they generally only use single-domain data rather than multi-source information and thus have limited performance. 

\subsection{Consumer Behavior Models}
Traditional consumer models \cite{huang2017label,qiu2015predicting,kooti2016portrait,ghazimatin2019fairy} only extracted shallow patterns from data to explain and improve models.
For example, Qiu et al. \cite{qiu2015predicting} designed a two-stage theory that uses a motivation factor and a product choice factor to simulate the consumer's decision-making process.
Kooti et al. \cite{kooti2016portrait} extracted the personal financial background information and combined them with demographic data to characterize and predict consumer behavior. They also studied the periodical behavior patterns, such as the intervals of two purchase and the frequency of shopping online, to make the model more explainable, and achieved excellent performance. However, the shallow patterns may be difficult to acquire the actual consumer mind patterns.
%
Some work focused on extracting high-level theoretical information from consumers' behaviors.
For example,
Joo et al. \cite{joo2015automated} extracted the hidden high-level data patterns--social traits \cite{abele2007agency}. According to the psychological theory, they utilized social traits to help them improve and explain how people make decisions in elections and estimate social relationships. Similarly, Kooti et al. \cite{su2018user} annotated the dataset with satisfaction and intent in searching and then designed the models to predict the possible feedback from the consumers to explain why machine learning chose the searching results. In consumer financial decision field, the business could provide professional domain theory for deep learning to model consumers. For example, Jisana et al. \cite{jisana2014consumer} introduced the frameworks of consumer models, such as the Maslow's hierarchy of needs~\cite{mcleod2007maslow} which described the deep inside need to explain the action of the purchase; AIO theory (Activities, Interests, and Opinions) \cite{boote1980psychographics} which depicted the daily life reason for purchasing an item. These consumption models could effectively extract consumer behavior patterns and catching their thoughts.

\subsection{GAN-related Work}
Generative models have been widely used for generating samples and domain alignment. In particular, GAN
has good flexibility and extensibility, and various factors and interactions can be incorporated into GAN.
Our work is related to Conditional Generative Adversarial Network (CGAN)~\cite{cgan}, which viewed the real samples as conditional constraints to guide data generation. Conditions could be any additional information, such as class labels and other modal data, which makes GAN applicable to cross-domain problems. Based on CGAN, Zhu et al. \cite{domainadaption1} studied the electronic signal shifting problem in the brain-machine interface. They used the GAN structure as the domain aligner to regularize the variational autoencoder.
Farshchian et al. utilized text to generate the corresponding images data to expand the sample space of certain classes, thus addressing class missing or imbalance.
Bousmalis et al. \cite{compare2} conducted the domain alignment by learning the picture style shift patterns under the unsupervised way, which could dismiss the limit of pairwise data. However, the above work failed to combine the non-pairwise data and pairwise data analysis in a unified network.
Our work leverage the above work in a unified structure, which exploits the advantages of business theory and deep learning models. We design two training batches to help our model conduct asymmetric 
domain alignment and survey analysis enhancement.

\section{Empirical Analysis}
We study the FTR prediction problem through the transaction and questionnaire information from an asymmetric two-source dataset.
Though it is challenging to utilize transaction data and questionnaire data effectively, in this section, we introduce the details of empirical analysis to extract high-level information which may relate to FTR based on business domain knowledge and reveal the bad distribution problem in questionnaires.
The dataset consists of 104,960 consumers' consumption activities in a month while only 4,492 of the consumers provide questionnaire feedback.
The consumers are divided into four classes according to FTR levels of their investment portfolios and we let class 0-3 represent low, medium-low, medium-high and high, respectively.
Table \ref{tab:data set Description} shows the basic statistics of the dataset.

\begin{table}
  \caption{Data Statistics
  }
  \label{tab:data set Description}
  \footnotesize
  \begin{tabular}{p{1.7cm}p{1.95cm}||p{1.8cm}p{1.9cm}}
  \toprule
  \multicolumn{2}{c|}{Questionnaire Information}&\multicolumn{2}{|c}{Consumer Consumption Information}\\
  \midrule
 \#Feedback & 4,492 &  \#Records & 8,297,231\\
 \#Question  & 52 & \#Consumer & 104,960\\
 Option & 1$-$7 &  AVG Expense & \$30,934\\
 \#FRT Class & 4 & AVG \#Trans & 79\\
 AVG Score & 5.43 &  Total Expense & \$18.98$-$\$1.01e+8\\
 \bottomrule
  \end{tabular}
\end{table}

\subsection{Consumption Activity Analysis}\label{sec:Data Character}

We divide consumers into ten groups in ascending order of their total personal expenses to discover consumption patterns based on the groups and investigate the overlap ratio of shopping scopes of two groups to deep mining the shopping behavior relationship between groups:
\begin{equation}
Overlap(a,b) = \frac{\#(a\cap b)}{\#(a\cup b)}
\end{equation}
where $a$ and $b$ denote two arbitrary groups' shopping scopes, and $\#$ denotes the element number in a set. 

\begin{figure}
    \centering 
    \subfigure[Consumption Capacity]{\includegraphics[width=0.24\textwidth,height=32mm]{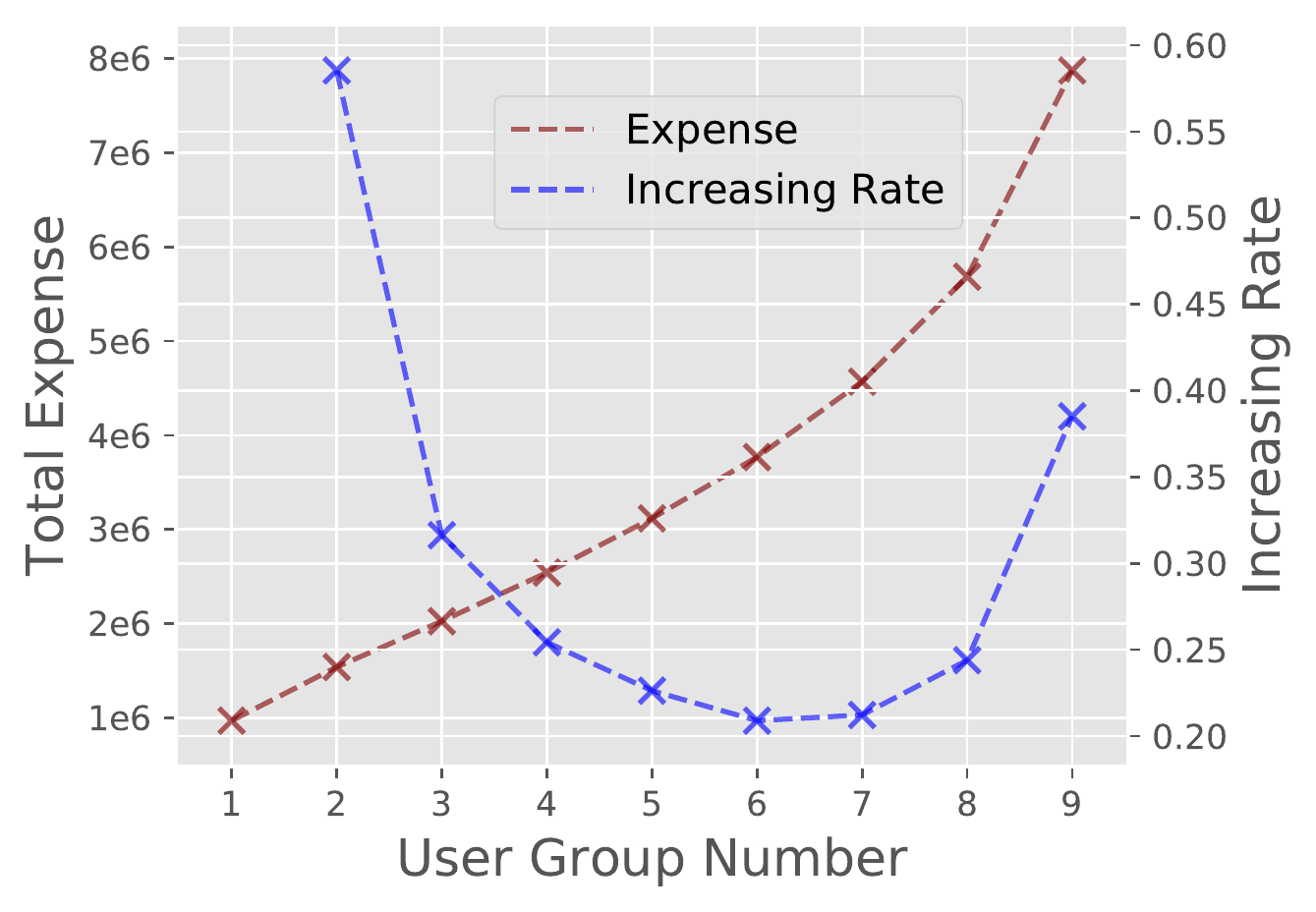}}
    \subfigure[Shopping Scope Overlap 
    Rate]{\includegraphics[width=0.24\textwidth]{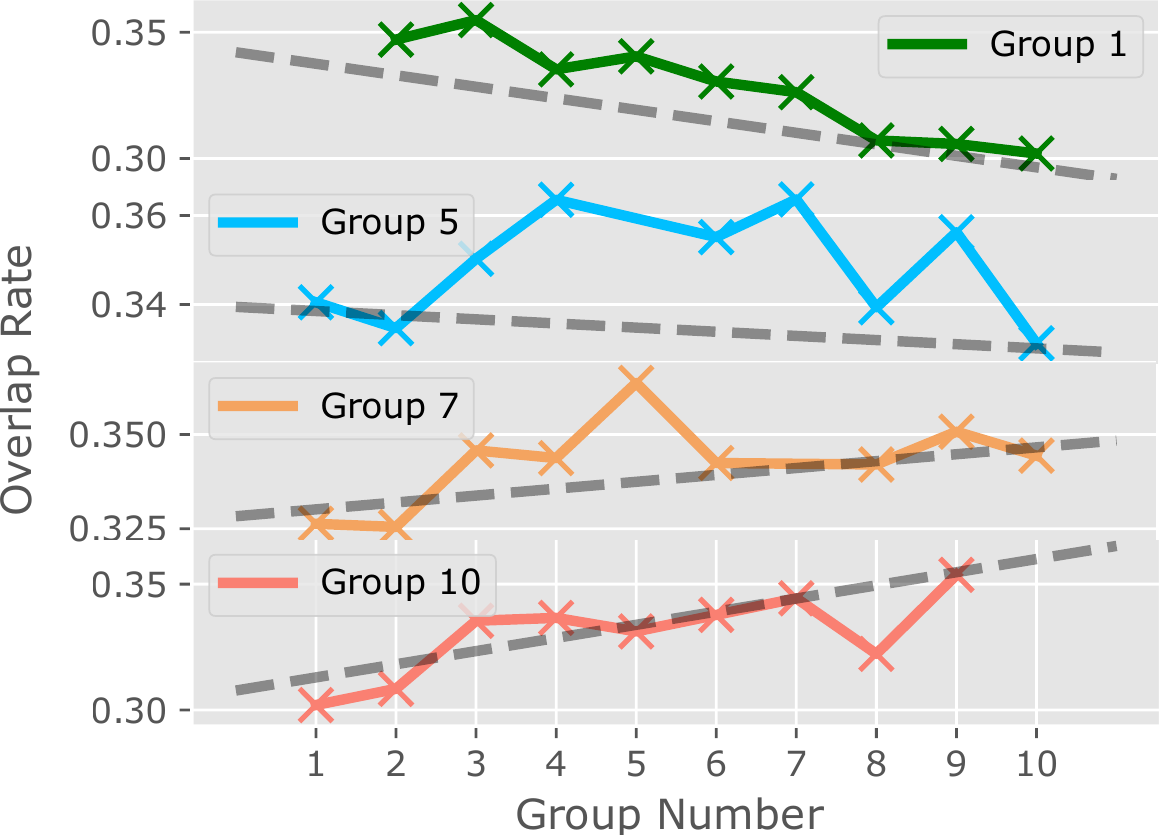}}
    \caption{(a) Illustration of consumption capacity. We omit Group 10, which is too large compared to other groups. Note Group 10 has the total expense of nearly \$19.12e+8 and a 5-time increasing rate. (b) Illustration of four representative groups' (group 1, 3, 7, and 10) pairwise overlap rates with other groups.}
    \label{tab:highestamount}
\end{figure}

Fig.~\ref{tab:highestamount}(a) exhibits the total expense grows, then stays stable, and finally soars from group 1 to group 9 as the increasing rate first drops, indicating large gaps between the consumption capacities of different consumer groups. Fig.~\ref{tab:highestamount}(b) provides further analysis with the relationship between the consumption capacity and shopping choices. The first few groups' overlap rate trends are descending, the middle groups' are stable, and the last few groups are ascending.


Based on the patterns in consumption capacity and shopping scopes, we conjecture three possible consumer categories:
\begin{itemize}
    \item \textit{Weak capacity consumers.} The first few groups only share more similar shopping choices with the consumers who have low expenses in the month.
    \item \textit{Ordinary capacity consumers.} The middle few groups share a relatively higher ratio of choices with all groups than other groups; they have more common with near groups.
    \item \textit{Strong capacity consumers.} The last few groups share more common with those who spend much money.
\end{itemize}

\subsection{Questionnaire Activity Analysis}\label{sec:survey_analysis}

The biopsychosocial questionnaire
investigates 4,492 consumers' feedback on 52 questions related to public concerns~\cite{lichtenstein2004effect,hartmann2012consumer}, collectivism-individualism~\cite{triandis2018individualism,chan2001toward}, and innovativeness~\cite{roehrich2004consumer,mandrik2005exploring}. All questions are answered on a 7-point scale, ranging from 1 (never or definitely no) and 7 (always or definitely yes).
%
Sample questions from each dimension are:
    `\textit{Is mankind severely abusing the environment?}',
    `\textit{Do I feel good when I cooperate with others?}', and
    `\textit{Do I know more than others on the latest new products?}'.

\begin{figure}[t]
    \centering 
    \subfigure[\quad Similar Consumer Feedback]{\includegraphics[width=0.24\textwidth]{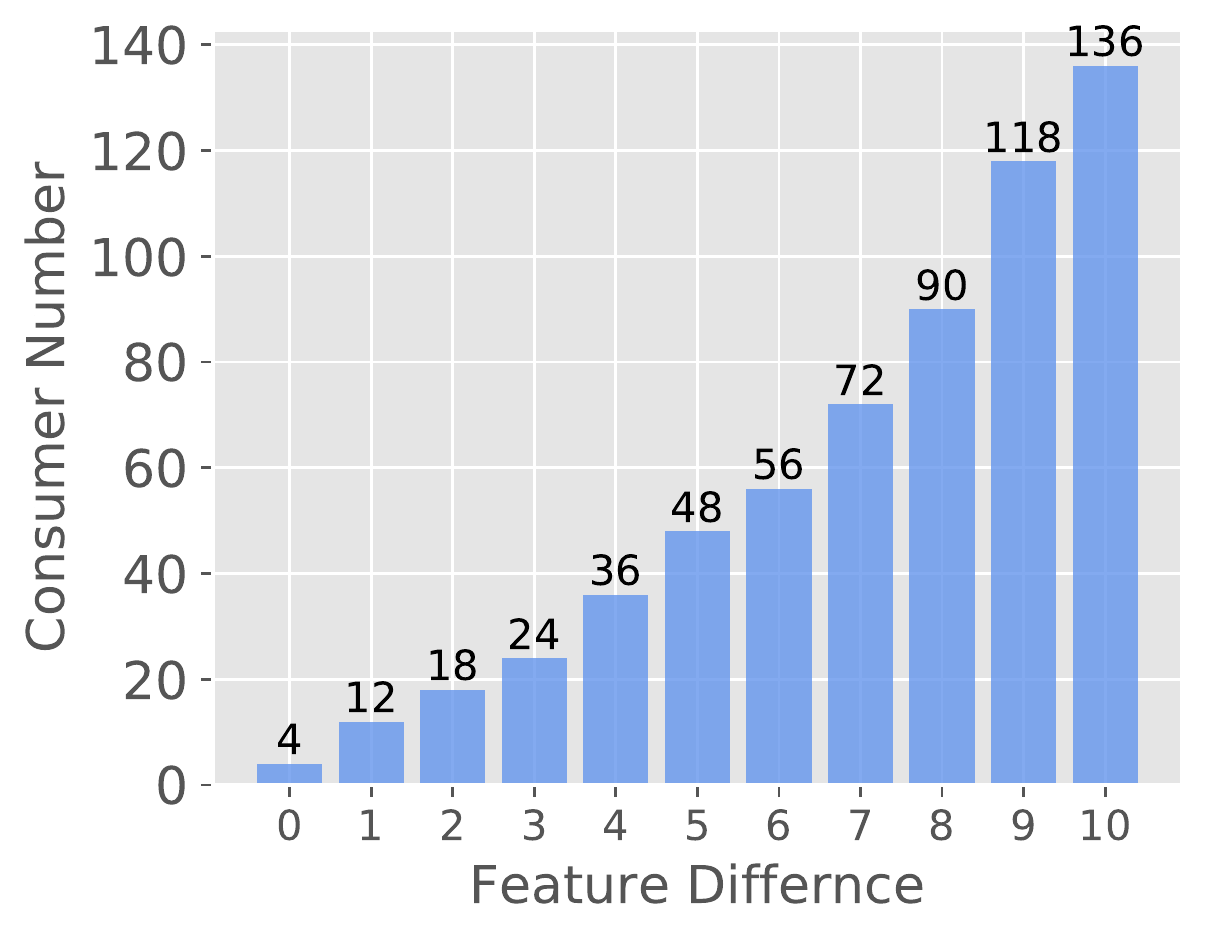}}
    \subfigure[\quad Feedback Distribution]{\includegraphics[width=0.24\textwidth]{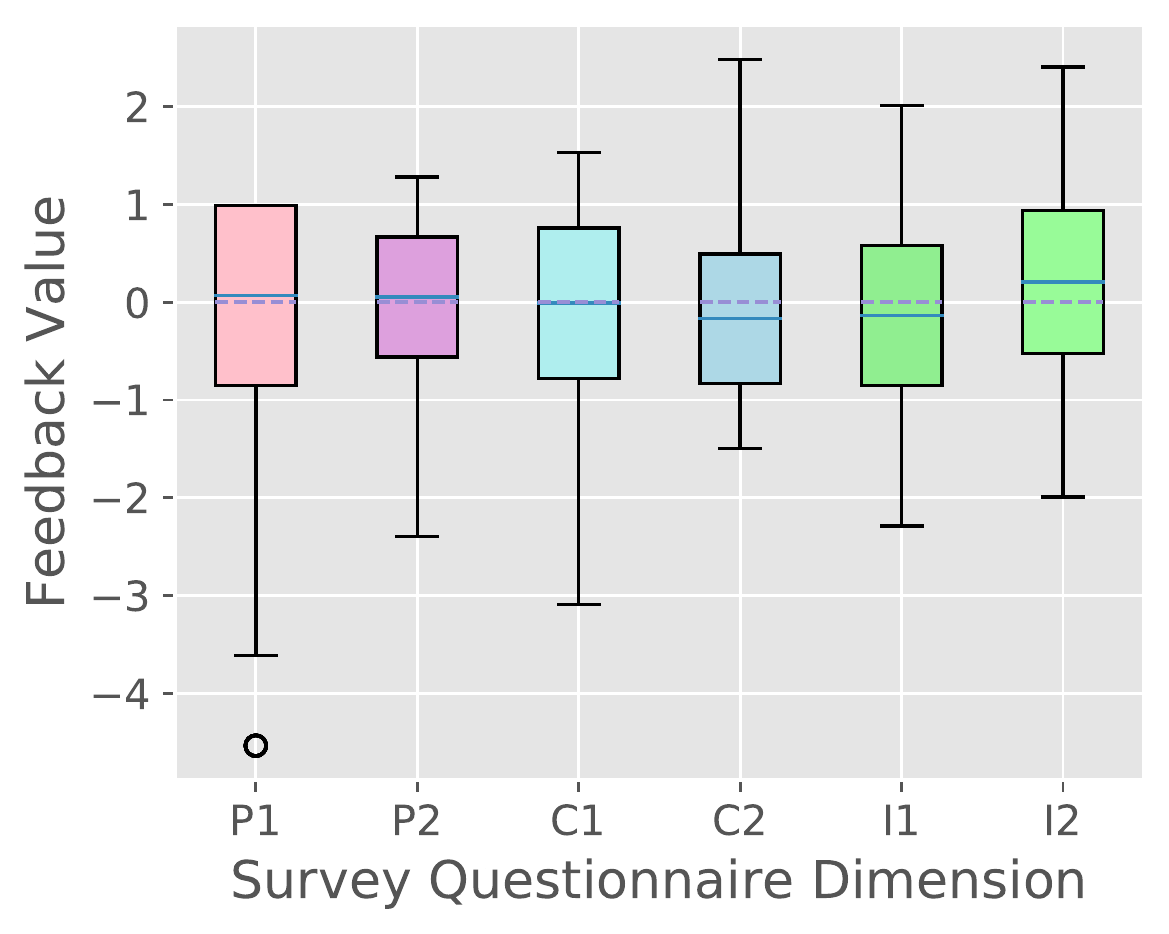}}
    \caption{(a) plots the multivalued function phenomenon in the questionnaire data. The number of consumers who have different FRT classes but give similar feedback in an accumulative way. (b) plots six features of questionnaires to show the limited scope in questions. We pick two questions from each dimension, name them P1, P2, C1, C2, I1, I2, respectively. The blue line and purple dashed line denote the median line and mean line of single feature distribution.}
    \label{fig:survey.eps}
\end{figure}

The consumers have an imbalanced distribution over four FRT classes---there are 700, 997, 2076, and 719 consumers in the four classes from low to high FRT.
Fig.~\ref{fig:survey.eps}(a) shows the number of consumers from different classes while they have little difference in the features. We have four consumers 
who have
the same features while they choose different portfolios from different FRT levels; near 50 consumers have different FRT while their feedbacks to 95\% of the investigated questions are similar, showing the multivalued function phenomenon in the data. Therefore, the data space is under-qualified, and we need to improve the data quality by using other domain information.

Fig.~\ref{fig:survey.eps}(b) shows the data distributions on several survey dimensions, where
most feedback is close to mean value (within $\pm1$ from the mean score). 
The survey data show an imbalanced class distribution where 46.2\% of the consumers belong to the third class, and
there exist many similar consumers from different classes.
The overlapping rate between two domains of data sources is only 4.3\%.




\subsection{Consumption Feature Representation}\label{sec:features}
\begin{table}
  \caption{Purchasing Features of Consumers}
  \label{tab:consumer Feature}
  \scriptsize
  \resizebox{\columnwidth}{!}{
  \begin{tabular}{ccccc}
  \toprule
  \textbf{BASIC} & \textbf{SOCIAL} & \textbf{SELF} & RESTAURANT & ENTERTAINMENT\\
    SERVICE & TRAVEL & SHOP & HEALTH & WORK\\
   CREDIT & HOME & DAILY & INVESTMENT & BILL\\
   GAMBLING & EDUCATION & CHARITY & FASHION & TAX\\
   \bottomrule
  \end{tabular}}
  \vspace{-0.3cm}
\end{table}
Based on the assumption of three consumer types in Section \ref{sec:Data Character}, we build \textit{Consumer Consumption Model} (Table~\ref{tab:consumer Feature}) to portray the consumer shopping capacity and intent. The Stratum Features are in bold font and the others are Life Features.

\vspace{1mm}\noindent\textbf{Stratum Features.}
We design \textit{Stratum Features} by dividing consumers' consumption behaviors according to the Maslow's hierarchy of needs \cite{mcleod2007maslow} to portray the relationship between shopping choices and consumption capacity: the human needs are hierarchical and higher-layer needs only occur when the lower-layer needs are satisfied.
Therefore, We sort and classify the transaction into three hierarchical categories to represent the shopping capacity from different layer needs:
\textit{BASIC}, the consumption to address the fundamental needs of living;
\textit{SOCIAL} for enjoyment and improving life quality; and
\textit{SELF}, for enriching the mental life.
We further summarize stratum features of consumer $i$ by
\begin{equation}
Score(y)_i = \frac{Expense(y)_i}{\sum_{Stratum}{Expense(y)_i}}
\end{equation}
where $y\in Stratum$ denotes a dimension of Stratum feature and $Expense_i$ is the total personal expense of consumer $i$ in the corresponding dimension. The Stratum Feature is decided by the ratio to the total personal expense.


We have the following findings from the above stratum feature scores, and group types: (i) The Maslow's need hierarchy could capture the different need structures of different shopping groups; (ii) The significant gaps (shown in Fig.~1) can be eliminated by using stratum features; (iii) Stratum features characterize the patterns in Fig.~1, indicated by the ordinary, weak, and strong purchasing-capacity groups.

\vspace{1mm}\noindent\textbf{Life Features.}
We design \textit{Life Features} to describe the shopping structure of consumers based on the AIO theory (Activities, Interests, and Opinions) \cite{boote1980psychographics}, which covers all the life aspects of purchases.
We look into consumers' shopping frequency and choices of targets beyond the three general types.
One example is that, generally, a small portion of consumers contribute to the majority of sales in a market (the ``20\%:80\%" rule \cite{plummer1974concept}).
Another example is that a company selling clothes will be more interested in consumers who frequently buy fashion products instead of consumers who only spend money at home.
Therefore, we similarly label the transaction into 17 subgroups based on the transaction categories and summarize the frequencies (Table~\ref{tab:consumer Feature}) of consumer $i$ by
\begin{equation}
Score(y)_i = \frac{Frequency(y)_i}{Max(\{Frequency(y)_i:i\in Consumers\})}
\end{equation}
where $y\in Life$ denotes a dimension of Stratum feature and $Frequency_i$ is the frequency of the consumer $i$ shopping in the corresponding dimension. The Life Feature is decided by the ratio to the max frequency of all consumers in the corresponding life aspect.
We draw the following conclusions from the above observations: 
(i) Life features indicate the structures of consumers' purchase types;
(ii) Life features reflect the shopping choice characteristics, overlapping choices among different groups, and shopping scopes of consumers.

\section{Methodology}\label{methodology}
To address the bad distribution problem posed by the survey in Section \ref{sec:survey_analysis} and difficulties in utilizing consumer activity data, we first propose a domain-specific consumer model to analyze consumers' consumption activities, followed by utilizing GAN and Gaussian Noise to improve the dataset quality and to ease the class imbalance and the multivalued function issues.

\subsection{Problem Description}
Before introducing the methodology, we first define our problem.
We denote the consumer entity by $E$ and the two related information domains as survey information $S^{s}\in \mathbb{R}^{+}$ and activity information $S^{a}=\{Activity(i): i\in E\}$, where $S^{s}$ consists of numeric feedback of questionnaires, $Activity(i)$ is the set of consumption activity records of consumer $i$. We split the entity $E$ into two groups according to their known information as $E_{l}$ and $E_{f}$: if consumer $i$ has provided both survey information and activity information, then $i\in E_{f}$; if consumer $i$ only provides the activity information, then $i\in E_{l}$. Our goal is to predict the FRT class $L$ for consumers. Our core approach aims at utilizing the different combination of consumers from $E_{f}$ and $E_{l}$ to asynchronously train our model to achieve the above characters.

\subsection{Model Formulation}\label{model_definition}
\begin{figure*}
\begin{center}
\includegraphics[width=0.85\textwidth]{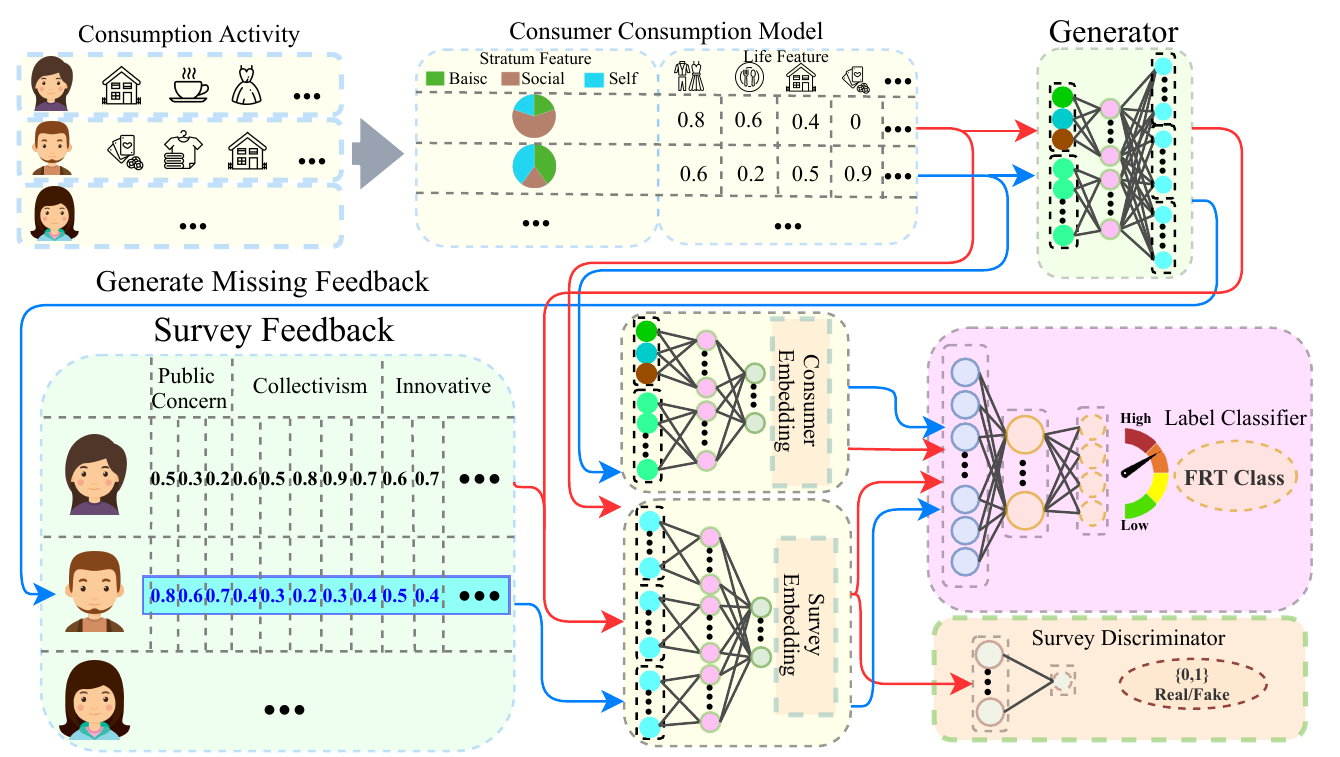}
\caption{Model overview. Our model takes the extracted consumer model information (Section \ref{sec:features}) and the survey information as inputs. We show how consumer data influence the modules in the model from $E_{l}$ and $E_{f}$ with blue and red lines, respectively. When training the Survey Discriminator, our model compares the generated survey information with the ground truth in the red path. When training Label Classifier and Generator, we take the generated survey as the 'real' features of consumers from $E_{l}$, and feed the model with the consumers from both $E_{l}$ and $E_{f}$.}
\label{tab:Model_architecture}
\end{center}
\end{figure*}
Our network Asymmetric Cross-Domain Generative Adversarial Network (ADGAN) takes consumer survey and consumption model information (Section \ref{sec:features}) as inputs.
ADGAN consists of three main structure: (1) Structural View Embedding, which effectively extracts the domain information, (2) Generator $G$, which generates the missing survey data, and (3) Discriminator $D$, which distinguishes fake from true samples and predicts the FRT labels. 

\vspace{1mm}\noindent\textbf{Structural View Embedding}. We design two types of domain-specific Structural View Embedding to conduct the domain alignment: Consumer Consumption Model Embedding and Survey View Embedding. We plot the schematic diagram in Fig.~\ref{tab:Model_architecture}. Since survey information $S^{s}$ and consumer consumption model $U$ can be divided into hierarchical structure according to the domain knowledge, it is intuitive to firstly extract the sub-dimension information, and then gather the dimension information to obtain the final embedding. For example, we will first analyze the public concerns, collectivism-individualism, and innovativeness, respectively, and then combine the analyzed information to measure the person's FRT.
Moreover, we will show through comparisons (Section \ref{Abalation Study}) that domain-specific embedding is effective and essential for extracting information from the provided data. Specifically, we apply a fully connected (FC) layer to extract information for each dimension and then combine the information using another FC layer. The FC layers in the Structural View Embedding will be followed by the activator--Rectified Linear Unit (ReLU).

\vspace{1mm}\noindent\textbf{Discriminator D}.
The structure of $D$ consists of two channels of inputs, and we use an FC layer activated by ReLU to receive the embedding information. We also set an FC layer with Sigmoid Function as the Survey Discriminator to distinguish the real and fake survey data, and an FC layer with Softmax Function as Label Classifier to predict the FRT classes. The structure and the training method of our $D$ are unique because we have two channels of inputs and two types of consumer groups.

Before introducing the loss function of our model, we introduce the batch construction.
We define two types of input combination (detailed in Section~\ref{sef:traning}):
First, given a consumer batch $B_{a}$, an arbitrary consumer $i\in B_{a}$ meets $\forall i\in E_{f}$---When we use the bath $B_{a}$, we have the pairwise survey ground truth and generated survey, in which case we can train the Discriminator how to distinguish fake data from true data; if the Discriminator can learn the patterns well, it can further help us improve the generated samples of consumers from $E_{l}$. 
Second, given a consumer batch $B_{b}$, there exist, two consumers, $i,j \in B_{b}$, $i\in E_{f}$ and $j\in E_{l}$---When we use the batch $B_{b}$, we have the ground truth data of the consumption activity, and FRT labels of consumers, in which case, if we have the well-trained Discriminator, we can improve our generator to interpret the consumer activity by providing the corresponding missed survey data under the condition of their FRT label and consumer model information; further, $E_{l}$ has more consumers from all labels, so the generated survey data can help ease the class imbalance problem of $E_{f}$.

Using the batch $B_{a}$, we have:
\begin{equation}\label{LD}
    \begin{aligned}
    L_{D}=\mathbb{E}_{U\sim\{U_{i}:i\in B_{a}\}}[D_{w}(G_{\theta}(U))]-\mathbb{E}_{S^{s}\sim\{S^{s}_{i}:i\in B_{a}\}}[D_{w}(S^{s})]\\+\lambda L_{GP}+\frac{1}{2}(L_{cls}(G_{\theta}(U)+L_{cls}(S^{s}))
    \end{aligned}
\end{equation}
where the first two terms calculate the Wasserstein distance of the distribution of real samples and fake samples; the third term $L_{GP}=\lambda(||\nabla_{\bar{S^{s}}}D_{w}(\bar{S^{s}})||_{2}-1)^{2}$ which refers to \cite{gradientpenalty} to improve the Wasserstein GAN training. The last two terms are the classification loss of fake and real samples, respectively.

Finally, we use $B_{b}$ to train a part of the Discriminator, $D_{align}$, and conduct the domain alignment.
$D_{align}$ includes a FC layer and classifier (the purple area in Fig.~\ref{tab:Model_architecture}) and is independent from the optimization of $D$.
We use $B_b$ to train $D_{align}$ and $G$; the loss function is defined in \eqref{GD}.

\vspace{1mm}\noindent\textbf{Generator G}. We train $G$ to generate the 'real' survey data from the corresponding consumer model information, which hopes to ease the FRT class imbalance in $E_{f}$ and helps interpret the consumer consumption behavior. Different from common CGAN, we only take the condition value $U$ as the input to $G$. Considering that we have more unknown samples than known samples (100,468 unknown and 4,492 known samples) and we will mix Gaussian noises with consumer activity information from $E_{f}$ (details in Section \ref{sef:traning}). We design $G$ with a Consumer Consumption Model Embedding and an FC layer activated by the Sigmoid function. The Sigmoid Function will help $G$ constrain the output scope between 0 and 1, which is similar to the scope of the compressed survey data in Section \ref{sec:survey_analysis}. Therefore, we have $\bar{S^{s}_{L_{i}}}\leftarrow{G_{\theta} (U_{i})}(i\in E)$, where $U_{i}$, $\bar{S^{s}_{L_{i}}}$ denote the consumer model information of consumer $i$, and generated survey information conditioned on the class label $L_{i}$, respectively. Then, we define the loss function of $G$ and $D_{align}$ as follows:
\begin{equation}\label{GD}
\resizebox{\columnwidth}{!}{
    $\begin{aligned}
    L_{G,D_{align}}=-\mathbb{E}_{U}[D_{w}(G_{\theta}(U))]+W*L_{cls}(G_{\theta}(U))+L_{cls}(S^{s}))
    \\s.t. \quad U\sim\{U_{i}:i\in B_{b}\cap E_{l}\},
    \quad S^{s}\sim\{S^{s}_i:i\in B_{b}\cap E_{f}\}
    \end{aligned}$}
\end{equation}

where the first term is the Wasserstein loss in \cite{arjovsky2017wasserstein}, and the other two terms are the weighted class constraint loss. We set $W_=\frac{|\{i:i\in E_{f}\cap B_{b}\}|}{|\{i:i\in E_{l}\cap B_{b}\}|}$ to prevent back-propagating over-weighted class loss of the generated samples.

We have the following findings from the above stratum feature scores and group types:

(i) The Maslow's need hierarchy could capture the different need structures of different shopping groups; 

(ii) The significant gaps (shown in Fig.~1) can be eliminated by using stratum features; 

(iii) Stratum features characterize the patterns in Fig.~1, indicated by the ordinary, weak, and strong purchasing-capacity groups.

\vspace{1mm}\noindent\textbf{Life Features.}
We design \textit{Life Features} to describe the shopping structure of consumers based on the AIO theory (Activities, Interests, and Opinions) \cite{boote1980psychographics}, which covers all the life aspects of purchases.
We look into consumers' shopping frequency and choices of targets beyond the three general types.
One example is that, generally, a small portion of consumers contribute to the majority of sales in a market (the ``20\%:80\%" rule \cite{plummer1974concept}).
Another example is that a company selling clothes will be more interested in consumers who frequently buy fashion products instead of consumers who only spend money at home.
Therefore, we similarly label the transaction into 17 subgroups based on the transaction categories, and summarize the frequencies (Table~\ref{tab:consumer Feature}) of consumer $i$ by
\begin{equation}
Score(y)_i = \frac{Frequency(y)_i}{Max(\{Frequency(y)_i:i\in Consumers\})}
\end{equation}
where $y\in Life, Frequency_i$ denote a dimension of Stratum feature and the frequency of the consumer $i$ shopping in the corresponding dimension. The Life Feature is decided by the ratio to the max frequency of all consumers in the corresponding life aspect.

\subsection{Fusion and Optimization}\label{sef:traning}
In this section, we will explain how we link consumer consumption model information and survey information in our Asymmetric Cross-Domain Generative Adversarial Network (ADGAN), and the technical details of batch construction methods driven by asymmetrical domain.
We provide the pseudocode of our overall training procedure in Algorithm \ref{algorithm}. In each epoch, we optimize $L_{D}$ in lines 2-10, where we mix the Gaussian Noise with $U$ to add diversity; we optimize $L_{G, D_{align}}$ in lines 11-18, where we mix the Gaussian Noise with $S^{s}$ to improve diversity. Note, we have three methods of sampling $B_{a}$ from $E_{f}$: undersampling, oversampling, and random. The sampling methods will influence the optimization of the discriminator, which will be compared in Section \ref{sec:comparison}. Due to the class imbalance in $E_{f}$, it is essential to ensure the balance between the sample spaces of classes. In this section, we will detailed explain how to split data for training from $B_{a},B_{b}$ (Definition in Section \ref{model_definition}) and Gaussian Noise fusion with $U,S^{s}$, respectively.

\vspace{1mm}\noindent\textbf{Batch $B_{a}$ Segmentation Method.}
We train the model abilities to deal with imbalance classes under random-sampling, oversampling, and undersampling, respectively. When random-sampling, we ignore the FRT class of consumers, and randomly select a batch of consumers from $E_{f}$; when oversampling, we sort consumers by labels, replicate the consumers until all classes have the same sample size, select a batch of consumers until exhausting all samples, shuffle and repeat the above options; when undersampling, similar to oversampling, but we randomly select samples from classes until a small size which is no higher than the fewest class number. After preparing the batch data for training, we need to improve the sample diversity before feeding batch data to the generator. Since we limit the supervised optimization of discriminator on the consumers from $E_{f}$, the consumer model information of them are stable. To make the discriminator and generator can apply in a more generalized situation, for each piece of consumer information, we independently sample several Gaussian noises from $\mathcal{N}(\mu,\sigma)$ and fuse them in random locations.

\vspace{1mm}\noindent\textbf{Batch $B_{b}$ Segmentation Method.} $B_{b}$ is designed for providing balanced samples and generating good biopsychosocial explanation for consumers from $E_{l}$. Since we have 100,469 unknown consumers and want to explain their activity, we will directly select some of them to supplement the real data without Gaussian Noise. We proportionally select consumers from $E_{f}$ by classes and supplement consumers from $E_{l}$ to make each class have the same training samples. Further, the consumer will have the same scores in their features because the questions in the survey only have seven options for choosing. Therefore, we similarly select and mix noise from $\mathcal{N}(\mu,\sigma)$ with the $\{S^{s}_{i}:i\in B_{b}\cap E_{f}\}$ to improve data quality.

\begin{algorithm}[h]
  \caption{Training procedure with default setting: oversampling the consumers for training discriminator, $I_{d}=20, \mu = 0,\sigma = 0.01,c=5, size=64$}
  \label{algorithm}
  \begin{algorithmic}[1]
   \Require the max epoch $Step$, batch size $size$, discriminator step number $I_{d}$, the hyperparameter for $L_{GP}$, Adam hyperparameters $\beta_{1}, \beta_{2}$, Adam learning rate $lr_{D},lr_{GD}$ for $L_{D}$ and $L_{G,D_{align}}$ Gaussian noise hyperparameters $\mu, \sigma$, Gaussian noise number $c$, sampling way for training discriminator.
    \For{$epoch \in [1,Step]$}
        \For{$iter \in [1,I_{d}]$}
            \State Oversample a batch $B_{a}$ of consumers from $E_{f}$ 
            \For{$i \in [1,size]$}
                \State Sample $c$ noise from $\mathcal{N}(\mu,\sigma)$ and mix with $U_{i}$ 
            \EndFor
        \State $\bar{S^{s}}\leftarrow{G_{\theta}(U)}$
        \State Compute the discriminator loss $L_D$ by \ref{LD}
        \State $w\leftarrow{Adam(\nabla_{w},w,lr_{D},\beta_{1},\beta_{2})}$
    \EndFor
    \State Sample a batch $B_{b}$ of consumers from $E_{f}$ and $E_{l}$
    \State $\bar{S^{s}}\leftarrow{G_{\theta}(U)}$
    \For{$i \in [1,size]$}
        \State Sample $c$ noise from $\mathcal{N}(\mu,\sigma)$ and mix with $S^{s}_{i}$
    \EndFor
    \State Compute the loss $L_{G,D_{align}}$ by \ref{GD}
    \State $w_{D_{align}}\leftarrow{Adam(\nabla_{w_{D_{align}}},w,lr_{GD},\beta_{1},\beta_{2})}$
    \State $\theta\leftarrow{Adam(\nabla_{\theta},\theta,lr_{GD},\beta_{1},\beta_{2})}$
    \EndFor
  \end{algorithmic}
\end{algorithm}

\section{Experiments}\label{sec:comparison}

\begin{table*}[!htb]
  \caption{
  }
  \label{tab:main experiment}
  \footnotesize
  \resizebox{\textwidth}{!}{
  \begin{tabular}{c|cccccccc}
  \toprule
  Model & 0-F1-Score & 1-F1-Score & 2-F1-Score& 3-F1-Score & Macro precision & Macro Recall & Macro F1-score & Accuracy \\
  \midrule
  LR & 0.01439(0.000)&    0.04255(0.000)&    0.42032(0.000)&    0.31214(0.000)&0.29814(0.000) & 0.29492(0.000) & 0.19735(0.000) & 0.29492(0.000) \\
  SVM & 0.00000(0.000)&    0.01550(0.000)&    0.41196(0.000)&    0.25610(0.000)&\textbf{0.46123(0.000)} & 0.28516(0.000) & 0.17089(0.000) & 0.28516(0.000)\\
  XGB & 0.02985(0.000)&    0.01429(0.000)&    0.41941(0.000)&    0.31214(0.000)&0.32154(0.000) & 0.29492(0.000) & 0.19392(0.000) & 0.29492(0.000) \\
\midrule
PixelDA(U) & 0.36000(0.120)&    0.02157(0.065)&    0.00448(0.013)&    0.03728(0.112)&    0.08203(0.059)&    0.25020(0.001)&    0.10583(0.017)&    0.25020    (0.001)\\
SGAN & 0.27555(0.083)&    \textbf{0.35137(0.068)}&    0.30023(0.070)&    0.04145(0.124)&    0.24669(0.031)&    0.29590(0.006)&    0.24215(0.010)&    0.29590(0.006)\\
GAZSL & 0.29600(0.019)&    0.10852(0.047)&    0.39152(0.013)&    0.44321(0.010)&    0.34468(0.014)&    0.34727(0.005)&    0.30992(0.013)&    0.34727(0.005) \\
PixelDA(S) & 0.38900(0.018)&    0.18701(0.039)&    0.39639(0.039)&    0.42766(0.035)&    0.35637(0.008)&    0.36328(0.004)&    0.34992(0.008)&    0.36328(0.004) \\
\midrule
RADGAN &0.00000(0.000)&    0.04957(0.070)&    0.44897(0.009)&    \textbf{0.49949(0.016)}& 0.25040(0.070) & 0.34941(0.006) & 0.24951(0.018) & 0.34941(0.006)\\
UADGAN &0.34417(0.021)&    0.15036(0.045)&    0.44413(0.022)&    0.48327(0.020)& 0.39792(0.015) & 0.38444(0.003) & \textbf{0.35548(0.013)} & 0.38444(0.003)\\
OADGAN &\textbf{0.39388(0.023)}&    0.03814(0.029)&    \textbf{0.48100(0.015)}&    0.49723(0.032)& 0.39252(0.080) & \textbf{0.40495(0.013)} & 0.35256(0.008) & \textbf{0.40495(0.013)}\\
 \bottomrule
  \end{tabular}}
\end{table*}

\subsection{Experiment Setting}\label{Experiment Setting}
Though our problem is unique and the dataset is private, it is difficult to find similar external datasets to help prove our methods. Therefore, we carry out comprehensive experiments on our dataset to show the effectiveness of our proposed methods.

We first use traditional classification algorithms as baselines: Logistic Regression (LR), Support Vector Machine , XGBoost (XGB)~\cite{chen2016xgboost}).
To form stronger baselines, we further selected three representative models from state-of-the-art cross-domain alignment GAN structures for the comparison 
and kept the best performance of different sampling methods as results:
GAZSL~\cite{gan1}, SGAN~\cite{odena2016semi}, PixelDA~\cite{compare2}.
The three models belong to supervised, semi-supervised, and unsupervised models, respectively.
We further derive supervised and unsupervised versions of PixelDA, denoted by PixelDA(S) and PixelDA(U).
Since these models are not originally designed for our dataset, we adapt them to our problem---we combine the auxiliary user model information in classifier module; the supervised models will only use the pairwise data; the semi-supervised and unsupervised models will utilize data and ignore the pairwise relationship.
The parameters of our methods are set as follows: $Step=4000$, $lr_{D}=1e-4$, $lr_{G}=1e-3$, $\beta_{1}=0.5$, $\beta_{2}=0.9$, $c=5$.
We run each model 10 times to obtain mean scores and standard deviations as experimental results.

We evaluate our model by comparing with several baselines and state-of-the-art methods, as well as conducting self-comparison. Table \ref{tab:main experiment} shows that baseline methods have extremely imbalanced performance in different classes. All three baselines perform poorly in predicting Class 0 and class 1. Although SVM achieves the highest Macro Precision score among all models, the baselines have the lowest Macro F1-score and Accuracy among the supervised methods due to the imbalanced performance. Compared to other GAN models, GAN theory largely eases class imbalance effects.

The unsupervised training method is largely weakened by the class imbalance. The semi-supervised SGAN performs best on Class 1, and worst output in Class 3, which means semi-supervised can better catch the Class 1 characters but will confuse the Class 3 training. The supervised methods, GAZSL and PixelDA(S), obtain the best performance while they only use the pairwise data. The supervised methods have balanced performance in all classes.
Comparison of our model with the supervised methods shows the non-pairwise data can provide extra information.

Among the sampling methods, randomsampling has the best performance in Class 3 but cannot distinguish Class 1 at all. Both undersampling and oversampling achieve better performance of our model than the other models.
While oversampling can extract the most information from the raw data, undersampling is more stable, demonstrated by a smaller standard deviation.

\subsection{Ablation Study}\label{Abalation Study}
We conduct ablation studies to evaluate the influence of View Structure Embedding and hyper-parameter settings.

\vspace{1mm}\noindent\textbf{View Structure Embedding.} To show the effectiveness of our Structural View Embedding and GAN, we set the comparison with the simple FC layer without view structure and the embedding structures in Fig.~\ref{tab:Model_architecture}: Survey FC layer (SFC), Consumer Consumption Model FC layer (CMFC), Combined FC layer (CFC); and View Structured Survey Embedding (SV), View Structured Consumer Consumption Model Embedding (CMV), and View Structured Combined Embedding (CV), where the combined analysis will concatenate the two domain data before feeding the output layer of the model. 
As shown in Table \ref{tab:View Structure}, all structural view embedding models outperform the corresponding FC layer in accuracy. We can clearly see the view structure improves the Precision and F1-Score in survey data analysis and combined data analysis, which show the effectiveness of the structural view embedding. 
The results also demonstrate that using combined data result in better overall representation better than using data from two domains independently.


\begin{table}
  \caption{Ablation study on mean (standard deviation) results.}
  \label{tab:View Structure}
  \footnotesize\resizebox{\columnwidth}{!}{
  \begin{tabular}{c|cccccccc}
  \toprule
   Model & Macro precision & Macro Recall & Macro F1-score & Accuracy \\
   \midrule
SFC &0.24186(0.029) & 0.28086(0.003) & 0.24343(0.014) & 0.28086(0.003)\\
  CMFC  & 0.32495(0.065) & 0.21953(0.001) & 0.18691(0.003) & 0.21953(0.001)\\
  CFC  & 0.27423(0.007) & 0.28184(0.005) & 0.27070(0.006) & 0.28184(0.005)\\
  \midrule
 SV  & 0.32603(0.018) & 0.32871(0.002) & \textbf{0.27701(0.007)} & 0.32871(0.002)\\
 CMV & 0.28559(0.074) & 0.25820(0.004) & 0.15523(0.032) & 0.25820(0.004)\\
 CV  &\textbf{0.34113(0.031)} & \textbf{0.33281(0.002)} & 0.26276(0.013) & \textbf{0.33281(0.002)}\\
  \bottomrule
  \end{tabular}}
\end{table}

\begin{figure}
    \centering
    \subfigure[Parameter - $c$]{\includegraphics[width=0.24\textwidth]{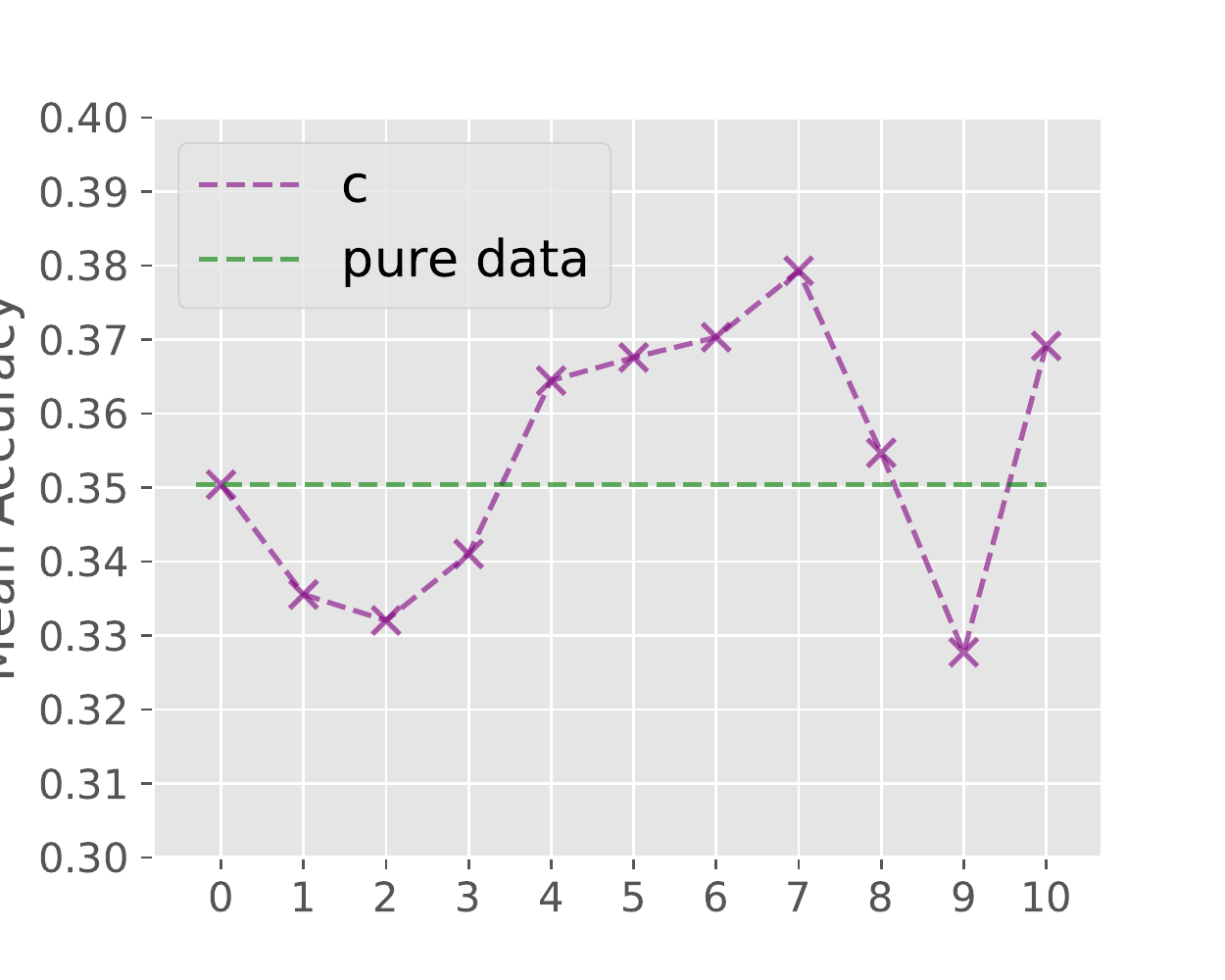}}
    \subfigure[Parameter - $lr$]{\includegraphics[width=0.24\textwidth]{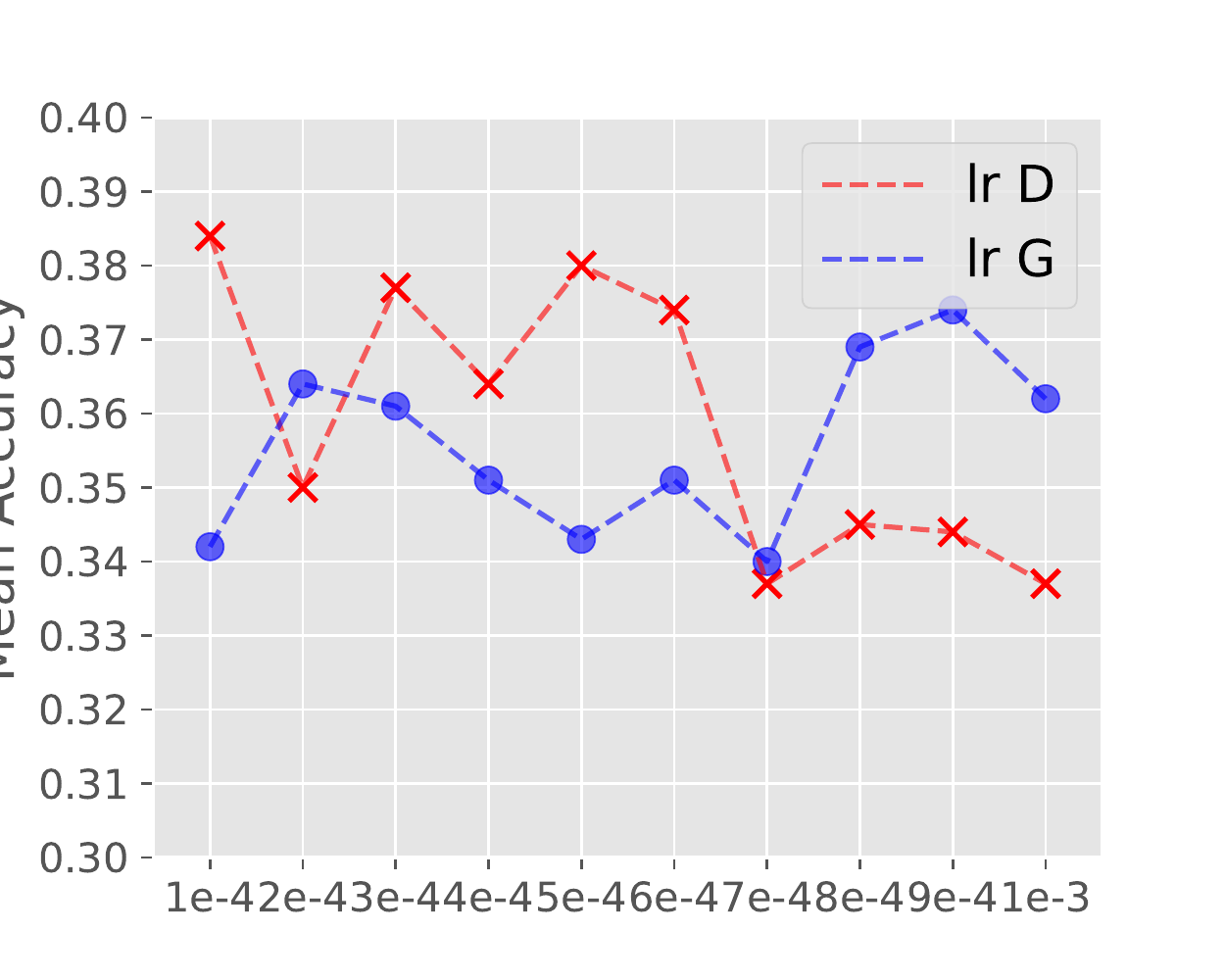}}
    \caption{(a) plots the mean accuracy of different Gaussian noise numbers in purple and the pure data baseline in green dashed lines; (b) plots the mean accuracy of the learning rates of Discriminator and Generator in blue and red lines.}
    \label{fig:ablation_on_learning_rate}
    \vspace{-0.3cm}
\end{figure}

\vspace{1mm}\noindent\textbf{Hyper-parameters.} We analyze how the Gaussian noise number $c$, and the learning rates $lr_{D}$, $lr_{G}$
influence the model performance, and prove the effectiveness of the Gaussian noise. We set undersampling, $Step=1000$, $lr_{D}=1e-3$, $lr_{G}=1e-3$, $\beta_{1}=0.5$, $\beta_{2}=0.9$, $c=5$ as defaulted values, and change the corresponding value when test the target parameter. We test the model for 5 times and conclude the mean performance. Fig.~\ref{fig:ablation_on_learning_rate}(a) and (b) plot the mean accuracy of models under different $c$, and $lr_{D},lr_{G}$, and we can see that the model obtains the best performance when $c=7$, $lr_{D}=1e-4$, $lr_{G}=9e-4$, respectively. The model will perform better with most Gaussian noise settings than the pure data; therefore, proper Gaussian Noise fusion can effectively improve the model performance. The mean accuracy and mean standard deviation of different learning rates for $D$ and $G$ are 0.35921 (0.017) and 0.35564 (0.011), respectively. The model is relatively stable under different learning rates, and we can conclude that the model will perform better if we set a relatively small $lr_{D}$ to $lr_{G}$.

\section{Conclusion}
In this paper, we construct domain-specific consumer models to describe consumers' consumption structures and intent. We further propose the Asymmetric Cross-Domain Generative Adversarial Network to extract information from unequal domains. our network outperforms state-of-the-art methods, and our ablation study reveals the influence of learning rates, as well as the effectiveness of structural view embedding and Gaussian Noise.
\bibliographystyle{IEEEtran}
\bibliography{sample-bibliography}
\end{document}